%% file: main.tex
\documentclass[letterpaper, 10 pt, conference]{ieeeconf}  

\IEEEoverridecommandlockouts                              

\usepackage{custom_symbols}
\usepackage{comment} 
\usepackage{float} 
\usepackage{cite}
\usepackage{graphicx} 

\title{\LARGE \bf Zero-shot Object-Centric Instruction Following: Integrating Foundation Models with Traditional Navigation}

\author{Sonia Raychaudhuri$^{1,2}$, Duy Ta$^{1}$, Katrina Ashton$^{1,3}$, Angel X. Chang$^{2}$, Jiuguang Wang$^{1}$, Bernadette Bucher$^{1,4}$
}

\begin{document}

\twocolumn[{%
\renewcommand\twocolumn[1][]{#1}%
\maketitle
\begin{center}
    \centering
    \captionsetup{type=figure}
\includegraphics[width=\linewidth]{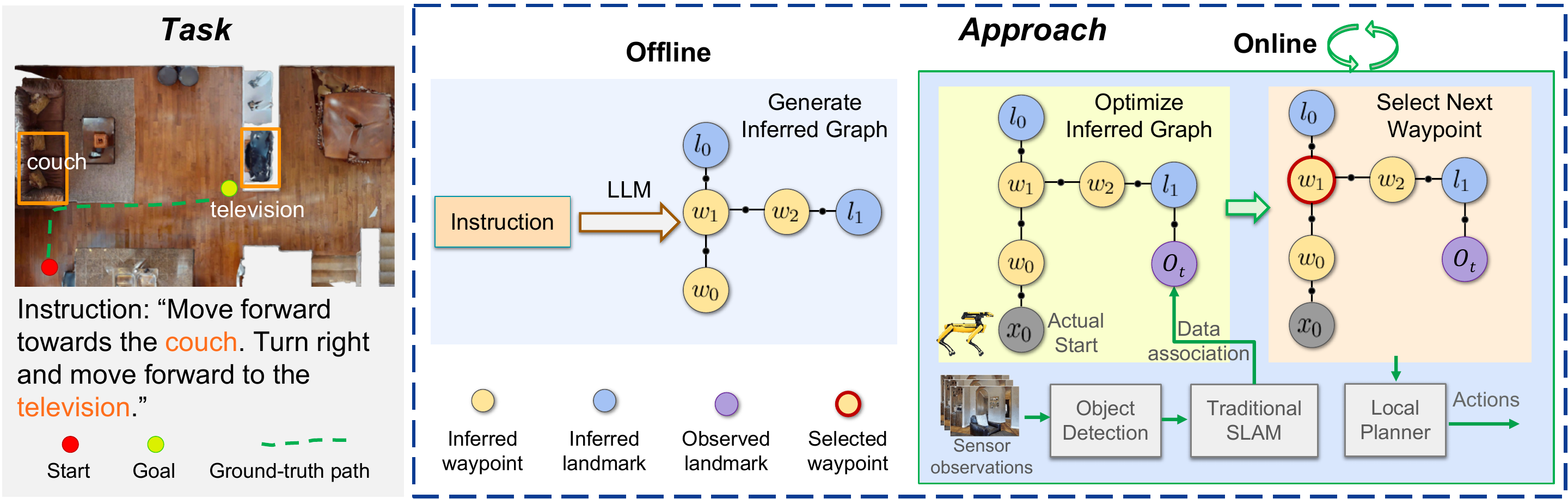}
\caption{
The robot is tasked with following a language instruction in an unseen environment to reach the final target (left). 
We introduce \methodnamefull (\methodname) (right), a novel method that infers a graph from the instruction using an LLM, offline, and thereafter optimizes the graph with actual observations at every step, online, during navigation. The robot selects the next waypoint from the inferred graph and moves toward it using a local planner.
}
\label{fig:teaser}
\end{center}%
}]

\input{chapters/abstract}

\footnotetext[1]{Work done during SR and KA's internships. SR, DT, KA, JW and BB are with the Robotics and AI Institute {\tt\small \{dta,jw\}@rai-inst.com}}
\footnotetext[2]{SR and AC are with Simon Fraser University {\tt\small \{sraychau,angelx\}@sfu.ca}}
\footnotetext[3]{KA is with University of Pennsylvania {\tt\small \{kashton\}@upenn.edu}}
\footnotetext[4]{BB is with University of Michigan {\tt\small \{bucherb\}@umich.edu}}%

\input{chapters/intro}
\input{chapters/related}

\input{chapters/dataset}
\input{chapters/method}
\input{chapters/experiments}

\input{chapters/results}
\input{chapters/conclusion}


\bibliographystyle{IEEEtran}
\bibliography{main}  


\end{document}

%% file: chapters/abstract.tex
\begin{abstract}
Large scale scenes such as multifloor homes can be robustly and efficiently mapped with a 3D graph of landmarks estimated jointly with robot poses in a factor graph, a technique commonly used in commercial robots such as drones and robot vacuums. In this work, we propose \methodnamefull (\methodname), a zero-shot method to ground natural language instructions in such a map. \methodname also includes a policy for following natural language navigation instructions in a novel environment while the map is constructed, enabling robust navigation performance in the physical world. 
To evaluate \methodname, we present a new dataset, \datasetname (\datasetnameshort), in order to evaluate grounding of object-centric natural language navigation instructions. 
We compare to two state-of-the-art zero-shot baselines from related tasks, Object Goal Navigation and Vision Language Navigation, to demonstrate that \methodname outperforms them across all our evaluation metrics on \datasetnameshort.
Finally, we successfully demonstrate the effectiveness of \methodname for performing zero-shot object-centric instruction following in the real world on a Boston Dynamics Spot robot.
\href{https://sonia-raychaudhuri.github.io/nlslam/}{sonia-raychaudhuri.github.io/nlslam}.

\end{abstract}

%% file: chapters/intro.tex
\section{Introduction}
\label{sec:intro}

Roboticists studying mapping and navigation have developed methods for creating landmark maps from visual observations and using those maps for autonomous navigation~\cite{thrun2006graph, mur2017orb, campos2021orb, rosinol2020kimera, rosinol2021kimera,labbe2019rtab,placed2023survey}. In fact, work in this area has matured far enough to move beyond the research community. Landmark-based mapping and navigation systems are widely used for autonomous robot operation in many industry applications, including home robot vacuum cleaners~\cite{robotvacuum,eade2010monocular} and drones for surveying and inspection~\cite{drones}. With recent advances in natural language understanding due to large language and vision-language models~\cite{openai2023gpt4, li2023blip2, radford2021learning,liu2023grounding,kirillov2023segment}, the next frontier is to determine how to accurately align natural language navigation instructions with observation-based autonomous decision making capabilities in these mature robotic systems. Effective alignment of this information could enable many new robotic capabilities including natural language instruction following.

To support research in this direction, researchers have developed benchmarks for evaluating an autonomous agent's ability to follow natural language navigation instructions in simulation, specifically R2R~\cite{anderson2018vision}, RxR~\cite{ku2020room}, and VLN-CE~\cite{krantz2020beyond}. These benchmarks consist of datasets of natural language instructions and associated paths through scanned Matterport 3D scenes of residential homes~\cite{chang2017matterport3d}.
The language instructions in these datasets contain both object-centric references to landmarks (`move forward to the \textit{blue chair}, turn left and stop at the \textit{wooden table}') and relative positional references to the current agent position (`move forward and turn left and stop in the hallway'). Methods addressing these tasks are largely trained on data in simulation and struggle to generalize to the real world~\cite{anderson2021sim,gervet2023navigating}.

\textbf{Contributions.} In this work, we introduce \methodnamefull (\methodname), a novel method for grounding natural language instructions to landmarks and robot poses in a factor graph, while performing zero-shot navigation in unseen unexplored environments. Our approach avoids previous limitations to performing instruction following robustly in the physical world by building off of robust mapping and navigation techniques for autonomous robots.
In addition, \methodname leverages pre-trained vision and language foundation models and requires no task-specific training. We demonstrate the viability of \methodname to successfully perform real-world instruction following on a Boston Dynamics Spot robot.

We are interested in achieving \textit{alignment} between natural language instructions and a widely-used landmark-based map structure in robotics as well as natural language navigation instruction following. To support both of these goals, we introduce a novel dataset, \textit{\datasetname} (\datasetnameshort), of object-centric navigation instructions. In this dataset, we ensure that the navigation instructions include references to landmarks which compose our sparse 3D map. We compare our method, \methodname, to two state-of-the-art zero-shot baselines from related tasks, Object Goal Navigation and Vision Language Navigation, to demonstrate that \methodname outperforms them across all our evaluation metrics on \datasetnameshort.

%% file: chapters/related.tex
\section{Related Work}
\label{sec:related}

\mypara{Natural Language Grounding to Landmarks.}
Recently there has been much interest in fusing information from 2D foundation models into 3D scene representations. Many approaches represent per-pixel semantic features using explicit representations such as pointclouds
~\cite{jatavallabhula2023conceptfusion,peng2023openscene,zhang2023clip,chen2023open}, or implicit neural representations~\cite{tschernezki2022neural,kobayashi2022decomposing,shafiullah2022clip,tsagkas2023vl,kerr2023lerf,huang2023audio,shen2023distilled,engelmann2024opennerf,mazur2023feature,qiu-hu-2024-geff,shi2024language,yamazaki2024open}. 
For methods which plan over objects instead of dense features, such scene representations necessitate extra storage and computation compared to object-centric representations. 
More similar to our approach, some methods~\cite{gu2024conceptgraphs,koch2024open3dsg,rosinol2021kimera} use a 3D scene graph (3DSG) where nodes and edges represent objects and inter-object relationships, respectively. Unlike us, however, these methods require constructing a dense 3D representation of the scene and some also do not consider pose uncertainty~\cite{gu2024conceptgraphs,koch2024open3dsg}. Also, unlike us, none of these approaches incorporate prior information about the scene. 



\mypara{Object Navigation.}
Navigation in unseen unstructured environments is a key challenge in robotics and AI. Research spans from simple point-to-point navigation (PointNav)~\cite{wijmansdd} to complex object navigation (ObjectNav) where a robot locates a target object specified by category (e.g. `picture')~\cite{batra2020objectnav} or a detailed description (e.g. `find the bottom picture that is next to the top of stairs on level one.')~\cite{qi2020reverie,zhu2021soon}. Some studies explore multi-hop navigation, requiring discovery of multiple objects using category~\cite{wani2020multion,raychaudhuri2024mopa} or multimodal cues~\cite{chang2023goat}. These tasks assess a robot's ability to find the target object through efficient and intelligent search of the environment by reasoning about semantic priors in the scene layouts~\cite{chaplot2020object,gervet2023navigating,yokoyama2023vlfm}. We instead focus on object-centric instruction following, where the robot receives dense instructions on how to reach the target (e.g. `move forward towards the cabinet then turn left and continue forwards until you reach the picture'). 
In this case the challenge is less efficient search, and more parsing the spatial relations of objects from complex language and grounding this structure to observations in the scene.

\mypara{Vision-Language Navigation.}
Vision-and-Language Navigation (VLN)~\cite{anderson2018vision,ku2020room,krantz2020beyond}, unlike ObjectNav, provides natural language instructions that specify both the target object and how to reach it (e.g. `head upstairs and turn left. stop in the hallway.'). 
This task challenges a robot to interpret detailed instructions, that combine actions (`turn', `walk'), spatial directives (`left', `up') and object or region references (`chair', `hallway’) in an unseen environment, making it much more challenging in terms of fine-grained language grounding.
While existing VLN datasets do not distinguish between object-centric (`move forward to the \textit{blue chair} and stop') and relative positional instructions (`move forward and stop'), we argue that these two instruction types require different reasoning capabilities and need to be studied independently. Our \datasetnameshort dataset focuses solely on object-centric instructions, enabling a more controlled study of object-centric instruction grounding in navigation.

A fundamental challenge in both VLN and \datasetnameshort is how to align dense natural language instructions to a path in the real world.  Trained methods can discover this alignment directly from image observations~\cite{majumdar2020improving,raychaudhuri2021language,krantz2021waypoint} or more commonly, through leveraging map-based approaches~\cite{chen2022think,wang2023gridmm,an2023bevbert,georgakis2022cross, liu2024volumetric}. 
However, these methods have struggled with sim-to-real transfer problems in the few cases where they have been run in the real world~\cite{anderson2021sim,gervet2023navigating}. 
Recently, much VLN research has focused on exploring the impact of large language models (LLMs) and vision-language models (VLMs) to solve VLN with zero-shot approaches
~\cite{dorbala2022clip, zhou2023navgpt, long2023discuss, chen2024mapgpt,zhan2024mc, huang2023visual,xu2023vision}.
Some zero-shot VLN methods even outperform trained approaches in the real world~\cite{long2023discuss,xu2023vision}. 
Some recent methods~\cite{chen2024mapgpt,zhan2024mc} construct topological maps 
and then use an LLM-based planner, thus relying entirely on an LLM for decision making while limiting the use of traditional navigation techniques from robotics. In contrast, we introduce a zero-shot method \methodname that uses an LLM not as a direct planner but instead to generate an inferred pose graph from the instruction. This serves as a prior which is gradually refined during navigation using well-established robotics techniques, allowing for a more structured and adaptive approach to instruction following.

%% file: chapters/dataset.tex
\section{Dataset}
\label{sec:dataset}

We introduce \textit{\datasetname} (\datasetnameshort) dataset for the instruction-following task, where a robot is required to navigate by following fine-grained object-centric instructions to reach a target. 
Unlike existing datasets, VLN~\cite{anderson2018vision} and VLN-CE~\cite{krantz2020beyond}, \datasetnameshort exclusively includes instructions with objects as landmarks, i.e. object-centric instructions (e.g., `move forward to the \textit{couch}, then go to the \textit{bench} in front of you and stop on the \textit{carpet}').


\mypara{Dataset Generation.}
We generate our episodes using real-world 3D scans from HM3DSem~\cite{yadav2023habitat} in the Habitat simulator~\cite{savva2019habitat}.
We utilize the Goat-Bench~\cite{khanna2024goat} dataset introduced for the GOAT task~\cite{chang2023goat} where the robot needs to navigate to a sequence of open-vocabulary objects specified via a mix of different modalities, i.e. category name, image or language description. In \datasetnameshort we use only one modality, i.e. language. But in contrast to the language description in Goat-Bench that describes only the target object, we generate fine-grained instructions that describe the entire route from the start to the target.
We do this by prompting an LLM (GPT-4~\cite{openai2023gpt4}) to generate navigation instructions from a series of RGB images along the route. We specifically prompt the LLM to include objects as landmarks in the instructions.
We do so in two stages -- \emph{Stage 1}: we prompt GPT-4 with the RGB images from the first and last frames of the trajectory and ask it to extract key objects; \emph{Stage 2}: we then prompt GPT-4 with all the sequential images collected for the trajectory along with the object names from Stage 1 and ask it to generate a fine-grained object-centric instruction.



\mypara{Statistics.}
\datasetnameshort contains episodes 
having open-vocabulary language instructions 
with 29 words, and 8 objects on average.
We support a continuous environment and an action space similar to VLN-CE (\Cref{tab:data_stats}). 
Although we have shorter path lengths, the instructions are still longer compared to VLN-CE, indicating that our instructions have more information (object-centric) for each path.

\input{tables/tab-dataset-statistics}

%% file: tables/tab-dataset-statistics.tex
\begin{table}[h]
\caption{
\textbf{Dataset comparison.}
Despite shorter path lengths, \datasetnameshort instructions are still long as they include object-centric information. 
}
\label{tab:data_stats}
\begin{center}
\resizebox{\linewidth}{!}{
\begin{tabular}{lccccc}
\toprule
Dataset & Environment & Action Space &Path Length & Actions & Instr. Len  \\
\hline
R2R &Discrete &Graph-based &10.0 &5 &29 \\
RxR &Discrete &Graph-based &14.9 &8 &129 \\
\midrule
VLN-CE &Continuous &2DoF & 11.1 &56 &19 \\
\datasetnameshort (Ours) &Continuous &2DoF &7.3 &44 &29\\

\bottomrule
\end{tabular}}
\end{center}
\vspace{-10pt}
\end{table}

%% file: chapters/method.tex
\section{Method}
\label{sec:method}

Our \methodname method stems from a key insight: language instructions not only guide the robot's navigation but also encode crucial spatial information about the environment's layout. Even before making any observations, these instructions provide the robot with a preliminary understanding of the environment's map, albeit with significant uncertainty. For instance, the instruction ``\textit{move forward until you see a chair.}" implies the presence of a chair somewhere along the forward direction (x-axis) relative to the robot's current position, even if we don't know the exact distance to the chair. By representing this spatial information as a factor graph~\cite{dellaert2017factor}, we can integrate it as a prior into a traditional factor-graph-based SLAM system. As the robot observes landmarks mentioned in the instructions, the uncertainty associated with these landmarks diminishes substantially, thus helping the robot in localizing itself within the context of the instructions during navigation. This effectively bridges the gap between linguistic guidance and spatial awareness. An overview of our method is shown in \Cref{fig:teaser}.

\subsection{Inferring map prior from language instructions}
\label{sec:llm_for_inferred_graph}

We transform language instructions into a \textit{language-inferred graph}, represented by a factor graph that encodes the prior distribution of the environment's map derived from the linguistic input. The graph includes two types of random variable nodes: (1) 
\emph{waypoint} nodes, $\waypoints=\{\waypoint_i \in \mathbf{SE}(2)\}$ representing the inferred, yet unknown, waypoints that the robot should navigate to, and
(2) \emph{landmark} nodes, $\inflandmarks=\{\inflandmark_j \in \mathbf{R}^2\}$ representing the unknown positions of objects that the robot is expected to observe along its path. 
For example, consider the instruction, ``\textit{go forward to the piano. turn right. stop at the table}". The resulting inferred graph, depicted in Fig. $\ref{fig:pose_optim}$, includes four waypoint nodes: the starting point $\waypoint_0$, the waypoint at the piano $\waypoint_1$, the right turn waypoint $\waypoint_2$, and the final stop at the table $\waypoint_3$. Additionally, there are two landmark nodes: $\inflandmark_0$ for the piano and $\inflandmark_1$ for the table.

The inferred graph represents the following joint distribution of waypoints and landmarks given their geometric relationships described in the instruction: 
\begin{equation}
  p(\waypoints,\inflandmarks|\wwlrelations) \propto p(\wwlrelations |\waypoints, \inflandmarks) = p(\wwrelations|\waypoints) p(\wlrelations|\waypoints,\inflandmarks)  \label{eq:inferred_graph}  
\end{equation} 
where $\wwlrelations=\wwrelations \cup \wlrelations$ is the set of all pre-defined geometric relations from the instruction, including inter-waypoint relations $\wwrelations=\{\wwrelation_{i,i+1}\}$ and waypoint-landmark relations $\wlrelations=\{\wlrelation_{ij}\}$.
The graph includes two types of factors: inter-waypoint factors $p(\wwrelations|\waypoints) = \prod_{i} p(\wwrelation_{i,i+1}|\waypoint_i, \waypoint_{i+1})$ and landmark-waypoint factors $p(\wlrelations|\waypoints,\inflandmarks) = \prod_{ij} p(\wlrelation_{ij} | \waypoint_i, \inflandmark_j)$.

Inter-waypoint factors $p(\wwrelation_{i,i+1}|\waypoint_i, \waypoint_{i+1})$ capture geometric relations between consecutive waypoints based on action verbs. For ``\textit{go forward}",
the relative pose $\wwrelation_{01}$ is modeled as a Gaussian in $\mathbf{SE}(2)$, with mean $\wwrelation_{01} = (x>0, y=0, \theta=0)$ indicating forward movement.
The $x$ component has high variance due to distance ambiguity, while $y$ and $\theta$ have low variances. This approach extends to other actions, with means and variances derived from corresponding verbs.

Landmark-waypoint factors $p(\wlrelation_{ij} | \waypoint_i, \inflandmark_j)$ represent spatial relationships between waypoints and objects mentioned in the instructions. In our example, waypoint $\waypoint_1$ and piano landmark $\inflandmark_0$ are modeled as being close: $p(\wlrelation_{10} | \waypoint_1,\inflandmark_0) \sim \mathcal{N}(\mathbf{0}, \Sigma)$, where zero mean indicates no offset and small $\Sigma$ in the diagonal elements represent low positional uncertainty.

\mypara{Instruction text to inferred graph.}
We leverage a Large Language Model (LLM) to transform free-form textual instructions into a \textit{language-inferred graph}. We use GPT-4~\cite{openai2023gpt4} to decode the instructions into an ordered sequence of waypoints, landmarks, and their relationships via this prompt:
    ``\textit{You are an expert in guiding a home navigation robot. The robot wants to follow a detailed language instruction to reach a target destination. To successfully complete this task, you will break down the input `instruction' to output a list of `waypoint', `landmark', `waypoint-to-waypoint transition actions' and `waypoint-to-landmark spatial relationship'.}''
We further provide examples to the LLM to perform in-context learning (ICL)~\cite{brown2020language,liu2021makes,wu2023self}. 
\Cref{fig:llm_example} shows example outputs.
Next we form the \textit{language-inferred graph} with the waypoint nodes ($\waypoint_i$) from the `waypoint' list, landmark nodes ($\inflandmark_i$) from the `landmark' list, inter-waypoint factors from the `waypoint-to-waypoint transition actions' and the landmark-waypoint factors from the `waypoint-to-landmark spatial relationship'. For both the factor types, we initialize the Gaussian with a mean displacement of 2 meters in $x$ and $y$. The initial values of the mean depend on the type of transition actions and landmark-waypoint relations.
\input{figures/fig-llm-example}

\subsection{SLAM with inferred map prior}
\label{sec:inferred_pose_optim}
\input{figures/fig-pose-optimization}

Our framework integrates the language-inferred map prior into a traditional object-based SLAM navigation system. The prior map initially has high uncertainty due to the inherent ambiguity in linguistic descriptions and the lack of direct sensory information. As the robot navigates, it simultaneously localizes itself within this uncertain map while refining the map through observations, gradually transforming the vague linguistic description into a precise spatial representation.

The primary objective is to find the Maximum A Posteriori (MAP) estimate of the joint distribution over the robot poses $\poses$, observed landmarks $\obslandmarks$, language-inferred waypoints $\waypoints$ and landmarks $\inflandmarks$, conditioned on all sensor observations $\sensors$ and the linguistic descriptions of relationships among waypoints and landmarks $\wwlrelations$. Formally, this is expressed as:
$$\poses^*, \obslandmarks^*, \waypoints^*, \inflandmarks^* = \argmax_{\poses, \obslandmarks, \waypoints, \inflandmarks} \ p(\poses, \obslandmarks, \waypoints, \inflandmarks |\ \sensors, \wwlrelations)$$
which can be decomposed into two components:
$$p(\poses, \obslandmarks, \waypoints, \inflandmarks |\ \sensors, \wwlrelations) = p(\poses, \obslandmarks |\ \sensors) \ \ p(\waypoints, \inflandmarks |\ \poses, \sensors, \wwlrelations)$$
The first term, $p(\poses, \obslandmarks |\thinspace  \sensors)$ is the traditional landmark-based SLAM problem, which can be efficiently solved using state-of-the-art SLAM algorithms~\cite{dellaert2017factor,placed2023survey,labbe2019rtab}. The second term, $p(\waypoints, \inflandmarks | \poses, \sensors, \wwlrelations)$, incorporates the language-inferred map:
\begin{equation}
p(\waypoints, \inflandmarks | \poses, \obslandmarks, \sensors, \wwlrelations) \propto p(\wwlrelations | \waypoints, \inflandmarks)  p(\sensors|\inflandmarks,\poses)
\label{eq:augmented_inferred_graph}    
\end{equation}
Here, $p(\wwlrelations | \waypoints, \inflandmarks)$ is our inferred factor graph in eq. $(\ref{eq:inferred_graph})$, and $p(\sensors|\inflandmarks,\poses)=\prod_{ik}p(\sensor_{jk} | \inflandmark_j, \pose_k)$ are the landmark-observation likelihood factors, where $\sensor_{jk}$ is the observation associated with the inferred landmark $\inflandmark_j$ when the robot is at pose $\pose_k$. These factors are essential for reducing uncertainty in the inferred landmarks, subsequently enhancing the accuracy and reducing the uncertainty of the associated waypoints. The inferred graph augmented with landmark-observation factors can be efficiently optimized using standard factor graph SLAM packages, such as GTSAM~\cite{gtsam}, allowing us to integrate established SLAM solutions with our novel language-informed priors with real sensory observations.

\subsection{Data association of inferred landmarks and  observations}
\label{subsec:data_assoc_landmarks}
To create the landmark-observation factors $p(\sensor_{jk}|\inflandmark_j, \pose_k)$ in eq. (\ref{eq:augmented_inferred_graph}), we must perform data association between the inferred landmarks and the observations. 
In this context, a landmark could have multiple candidate observation matches, and one observation could potentially match multiple landmarks. 
While this data association problem could be solved using Expectation-Maximization (EM)~\cite{dellaert2000feature,bowman2017probabilistic} or hybrid inference on a discrete-continuous factor graph \cite{doherty2022discrete}, we opt for a simpler approach. 
Our method matches landmarks with observations pairwise, selecting the best match with cosine similarity between their CLIP~\cite{radford2021learning} text features: 
\begin{equation}
\inflandmark^* = \argmax_{\inflandmark} [\simcos(\mathcal{F}_t(\inflandmark), \mathcal{F}_o(\sensor))]
\end{equation}
where $\mathcal{F}_t(.)$ denotes the landmark features in the inferred graph, and $\mathcal{F}_o(.)$ denotes the observation features. If the similarity is below a threshold no association is made.
This simple approach is effective and computationally efficient, future work may explore more sophisticated data association techniques to resolve ambiguity better.

\subsection{Navigation policy}

Localizing the robot within the language-inferred graph enables us to use a straightforward navigation policy. The robot navigates by sequentially moving to inferred waypoint nodes in the graph, transitioning when within 0.5 meter of each, and stopping at the last node. However, high uncertainty in waypoint locations can lead to inaccurate navigation. In these cases, the robot should switch to exploration mode, seeking landmarks to reduce uncertainty about upcoming waypoints.
To balance exploration and exploitation, we employ the following strategy for waypoint selection:
\begin{equation}
\label{eq:waypoint_selection}
\waypoint^* = \argmin_{\waypoint_j \forall j} \big[
||\waypoint_j \ominus \pose_i||^2
+ \alpha\trace(\Lambda_{\waypoint_j})
\big]
\end{equation}
This equation selects the inferred waypoint $\waypoint_j$ that minimizes two factors: the distance between the waypoint and the current robot pose ($||\waypoint_j \ominus \pose_i||^2$), and the uncertainty of the waypoint, represented by the trace of its information matrix $\Lambda = \Sigma^{-1}$. 
The balance between these factors is controlled by the constant $\alpha$.

To determine the precise navigation target, we further sample a point from the posterior distribution $p(\waypoint^*)$ within the navigable region around $\waypoint^*$.  Initially, the posterior distributions of waypoints have high uncertainty, promoting exploration. As the robot gathers observations, uncertainty decreases, thus refining navigation targets. 

\input{figures/fig-method-fgraph-example}

%% file: figures/fig-llm-example.tex
\begin{figure}[ht]
\centering
\includegraphics[width=\linewidth]{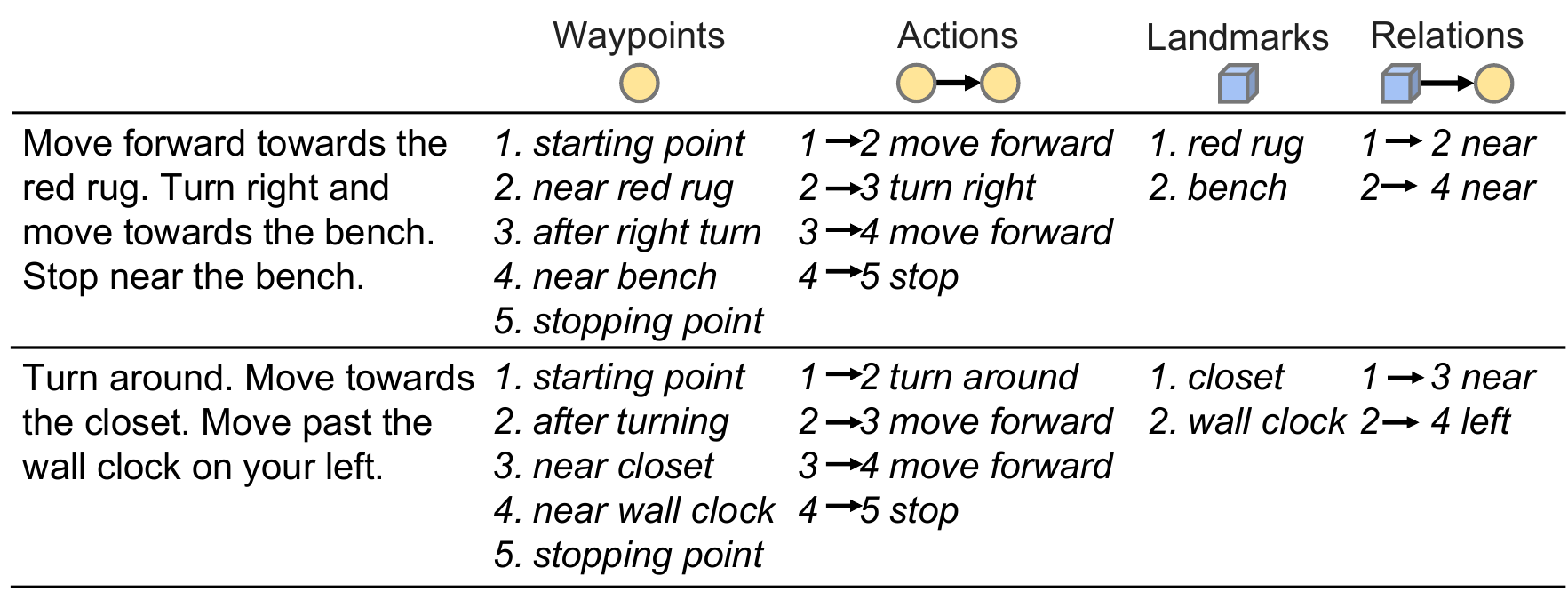}
\caption{
Using an LLM we extract waypoints, landmarks, waypoint-waypoint actions and landmark-waypoint relations.
}
\label{fig:llm_example}
\vspace{-5pt}
\end{figure}

%% file: figures/fig-pose-optimization.tex
\begin{figure}[ht]
\centering
\includegraphics[width=\linewidth,
]{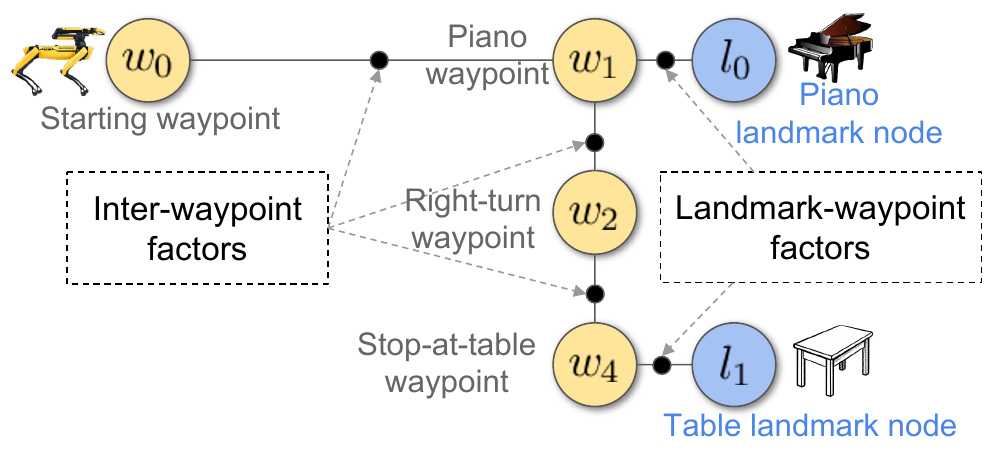}
\caption{
\textbf{Language-inferred factor graph} corresponding to the language instruction \textit{``Go forward to the piano. Turn right. Stop at the table"}. 
}
\label{fig:pose_optim}
\vspace{-10pt}
\end{figure}

%% file: chapters/experiments.tex
\section{Experiments}
\label{sec:experiments}
\mypara{Task.}
At each time step, the robot has access to RGB and depth images, as well as the entire language instruction and is expected to output one of four actions: \textit{move forward}, \textit{turn left}, \textit{turn right} and \textit{stop}. We use the same agent embodiment and task setting as VLN-CE~\cite{krantz2020beyond}.

\mypara{Metrics.}
We use standard evaluation metrics for VLN~\cite{anderson2018vision,anderson2018evaluation,ilharco2019general}: Success Rate (SR), Success weighted by inverse Path Length (SPL), Oracle Success Rate (OSR), and normalized Dynamic-Time Warping (nDTW). The first three measure whether the robot can reach the target object, while nDTW measures how closely it can follow the instructions.

\mypara{Implementation.}
We evaluate \methodname on \datasetnameshort in the Habitat simulator~\cite{savva2019habitat} in a zero-shot manner.
We use a pipeline of pre-trained methods for landmark recognition.  We use RAM~\cite{zhang2024recognize} to identify objects, followed by Grounding-Dino~\cite{liu2023grounding} to predict bounding boxes, and finally, Segment-Anything (SAM)~\cite{kirillov2023segment} to obtain the semantic segmentation. For the navigator, we use a PointNav policy~\cite{wijmansdd} pre-trained on HM3D~\cite{ramakrishnan2021hm3d} scenes.
We compare to two zero-shot baselines, a SOTA ObjectNav method adapted to our task and an LLM-based SOTA method for instruction-following:

\myparawindent{\baselinemethod baseline.}
VLFM~\cite{yokoyama2023vlfm} is a single-object navigation method, which we execute sequentially for all objects, extracted using GPT-4~\cite{openai2023gpt4}, from instructions. 
Using BLIP-2~\cite{li2023blip}, VLFM creates value maps of likely object locations, guiding the frontier-based exploration~\cite{yamauchi1997frontier} to efficiently find objects. Note that this baseline completely disregards fine-grained navigation instructions on how to reach the target.

\myparawindent{\instructnavbaseline baseline.} InstructNav~\cite{long2024instructnav} is a method for general navigation instruction following which evaluates on Object-Goal navigation, VLN, and demand-driven navigation tasks. InstructNav uses an LLM to break navigation instructions into a series of paired actions and landmarks. It then plans with an LLM on how to reach each landmark using a combination of value maps representing key information about the environment 
such as semantic information, actions, trajectory history, and intuition. 
While InstructNav uses LLM as a planner, our method efficiently uses LLM offline to generate inferred graphs from instructions just once.

%% file: chapters/results.tex
\section{Results}
\label{sec:results}

\mypara{Benchmark results.}
\Cref{tab:results_main} shows that \methodname outperforms the baseline methods on all metrics. 
Our performance on the nDTW metric (48\%) specifically indicates that our method is able to follow instructions better than the others.
This is intuitive since our method transforms the instructions into a graph, which is then used to perform the navigation, compared to \baselinemethod, which searches for objects mentioned in the instructions by disregarding the actions and relations encoded into them, and InstructNav, which uses LLM to directly plan over the instructions.
Being able to follow the instructions also enables \methodname to successfully reach the target object as indicated by its performance on the SR and SPL metrics.
The OSR metric reflects whether the agent has been near the target object but failed to stop. We find that all the methods achieve reasonable OSR, indicating that the objects are easy to navigate to. However, the gap between SR and OSR is smaller for \methodname (20\%) compared to the others (42\% in \baselinemethod and 43\% in InstructNav), indicating that our method is able to call the \textit{Stop} action correctly more often. 
This shows that the simple and straightforward stopping criteria in \methodname works fairly well, although a more sophisticated stopping policy may yield improvement.

\input{tables/tab-results}

\mypara{Qualitative analysis.}
In \Cref{fig:qual_rollout}, we visualize our agent performance on a single episode over time. At the beginning of the episode ($t=1$), when the agent has not yet made observations in the world, the landmark nodes, $\inflandmark_i=\{\inflandmark_1=``brown~couch", \inflandmark_2=``fireplace"\}$ and the waypoint nodes, $\waypoint_i=\{\waypoint_2,\waypoint_3\}$ have large uncertainties (visualized as ellipses). As the episode progresses, the agent makes observation and grounds $\inflandmark_1$ 
and $\waypoint_1$, thereby becoming certain about their locations. At $t=50$, the agent observes $\inflandmark_2$, thus reducing uncertainty for all the nodes in the inferred graph. Finally, the agent generates the \textit{Stop} action when it is within 0.5m of the last waypoint $\waypoint_2$ in the graph.
\input{figures/fig-qualitative-episode-rollout}
To compare \methodname and \baselinemethod, we visualize the agent trajectories for different episodes (\Cref{fig:qual_comparison}) and observe that \methodname follows the instruction better compared to \baselinemethod. This is evident from how closely the agent path matches the ground-truth path in our method. 

\input{figures/fig-qualitative-comparison}

\mypara{Ablations.}
We perform ablations on the different modules in \methodname (\Cref{tab:ablations_main}) to understand the contribution of each. For both \textit{Object Detector} and \textit{Navigator}, we find using an Oracle performs the best, as is expected. The Oracle detector+navigator achieves a +12\% increase in success, +13\% increase in SPL, and +10\% increase in nDTW, indicating that the performance of our method is limited by the choice of the detector and the navigator and can be improved further.
\input{tables/tab-ablations}

\mypara{Failure analysis.}
We note two main failure cases: (a) the presence of multiple instances of the same object category; and (b) the misclassification of an object. 
In the first case, an instruction `move through an open door' in the presence of multiple doors (\Cref{fig:error_analysis}) makes it difficult for the agent to figure out the correct `door'. 
The second case is a limitation of the object detector where it incorrectly identifies a `television' as `fireplace', leading to erroneous data association.

\input{figures/fig-error-analysis}

\mypara{Real-world demonstrations.}
We deployed \methodname on a Spot robot from Boston Dynamics to demonstrate its effectiveness in the real world. We replaced the PointNav policy with the Boston Dynamics Spot SDK as the navigator to move the robot to a waypoint. We use RTAB-Map to obtain the robot's coordinates and the observations in a reference frame. Instead of our object detection pipeline, we use YOLOv7~\cite{wang2023yolov7} for faster real-time execution. We ran \methodname in an office space with commonly occurring objects such as chairs, tables, potted plants, etc., and demonstrated that our method was able to perform well in this task.

%% file: tables/tab-results.tex
\begin{table}[h]
\caption{
\textbf{Performance.} \methodname significantly outperforms  \baselinemethod and \instructnavbaseline on all metrics.
}
\label{tab:results_main}
\begin{center}
\begin{tabular}{lrrrr}
\toprule
Method & SR & SPL & OSR &nDTW \\
\hline
\baselinemethod & 0.33 &0.24 &0.75 &0.33 \\

\instructnavbaseline & 0.24 & 0.07 & 0.67 & 0.33 \\

\methodname (Ours) & \textbf{0.58} &\textbf{0.40} &\textbf{0.78} &\textbf{0.48} \\

\bottomrule
\end{tabular}
\end{center}
\vspace{-10pt}
\end{table}

%% file: figures/fig-qualitative-episode-rollout.tex
\begin{figure}[ht]
\centering
\includegraphics[width=\linewidth]{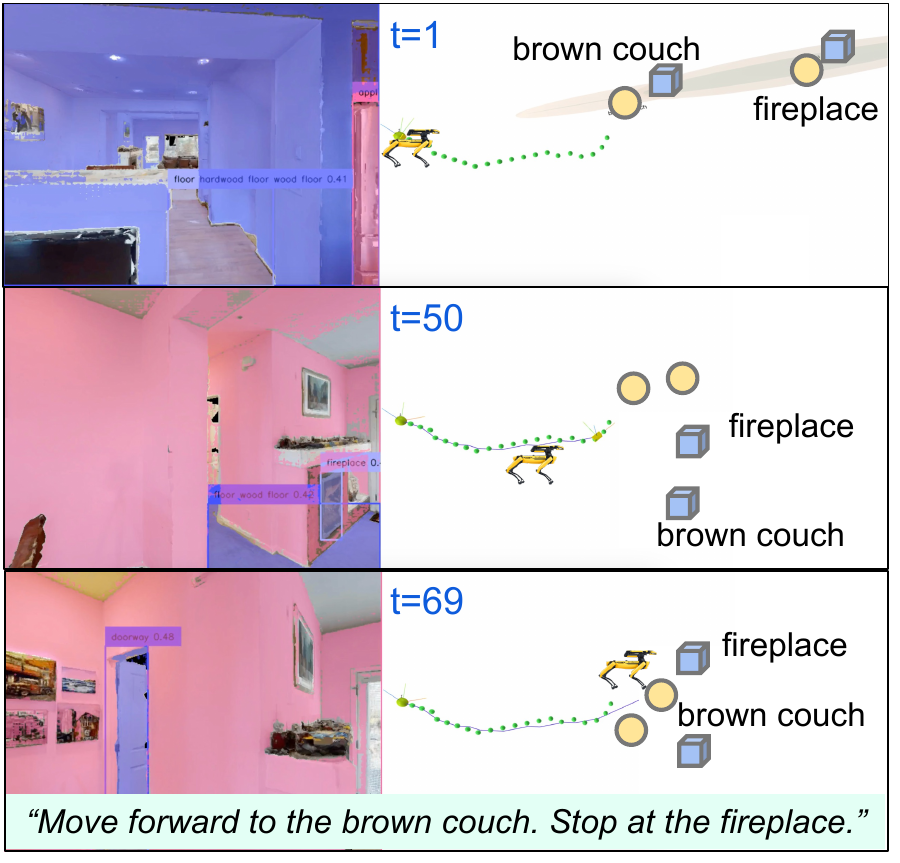}
\caption{
\textbf{\methodname in action.}
Visualizing the progress of our agent through an episode shows how the \textit{language-inferred graph} gets optimized over time $t$ by making observations in the real-world and performing data association, leading to a successful completion.
(left) shows detected objects; (right) shows current robot pose, ground-truth trajectory (green), inferred waypoints (yellow) and landmarks (blue). 
}
\label{fig:qual_rollout}
\vspace{-10pt}
\end{figure}

%% file: figures/fig-qualitative-comparison.tex
\begin{figure}[ht]
\centering
\includegraphics[width=\linewidth]{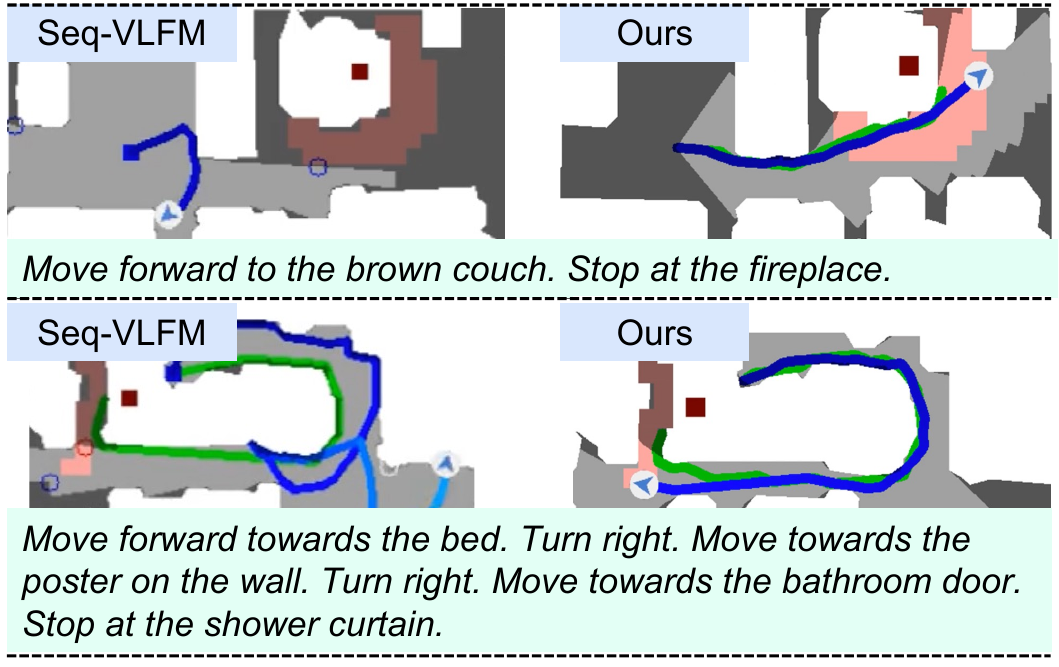}
\caption{
\textbf{Comparison.}
Our agent (right) trajectory aligns with the ground-truth trajectory better than  \baselinemethod (left), indicating better instruction following ability.
Agent trajectory is \textcolor{blue}{blue}, ground-truth path is \textcolor{green}{green} and target object is \textcolor{red}{red}.
}
\label{fig:qual_comparison}
\vspace{-10pt}
\end{figure}

%% file: tables/tab-ablations.tex
\begin{table}[h]
\caption{
\textbf{Ablations.} Oracle versions of the \textit{Object Detector} and the \textit{Navigator} perform better than the pre-trained models, indicating the scope of improvement for \methodname.
}
\label{tab:ablations_main}
\begin{center}
\resizebox{\linewidth}{!}{
\begin{tabular}{cc rrrr}
\toprule
Object Detector & Navigator & SR & SPL & OSR &nDTW \\
\midrule
Oracle & Oracle &\textbf{0.70}  & \textbf{0.53} & \textbf{0.80} & \textbf{0.58} \\
RAM+GDino+SAM & Oracle & 0.67 &0.48 &\textbf{0.80} &0.53 \\
RAM+GDino+SAM & PointNav & 0.58 &0.40 &0.78 &0.48 \\
\bottomrule
\end{tabular}
}
\end{center}
\vspace{-10pt}
\end{table}

%% file: figures/fig-error-analysis.tex
\begin{figure}[ht]
\centering
\includegraphics[width=\linewidth]{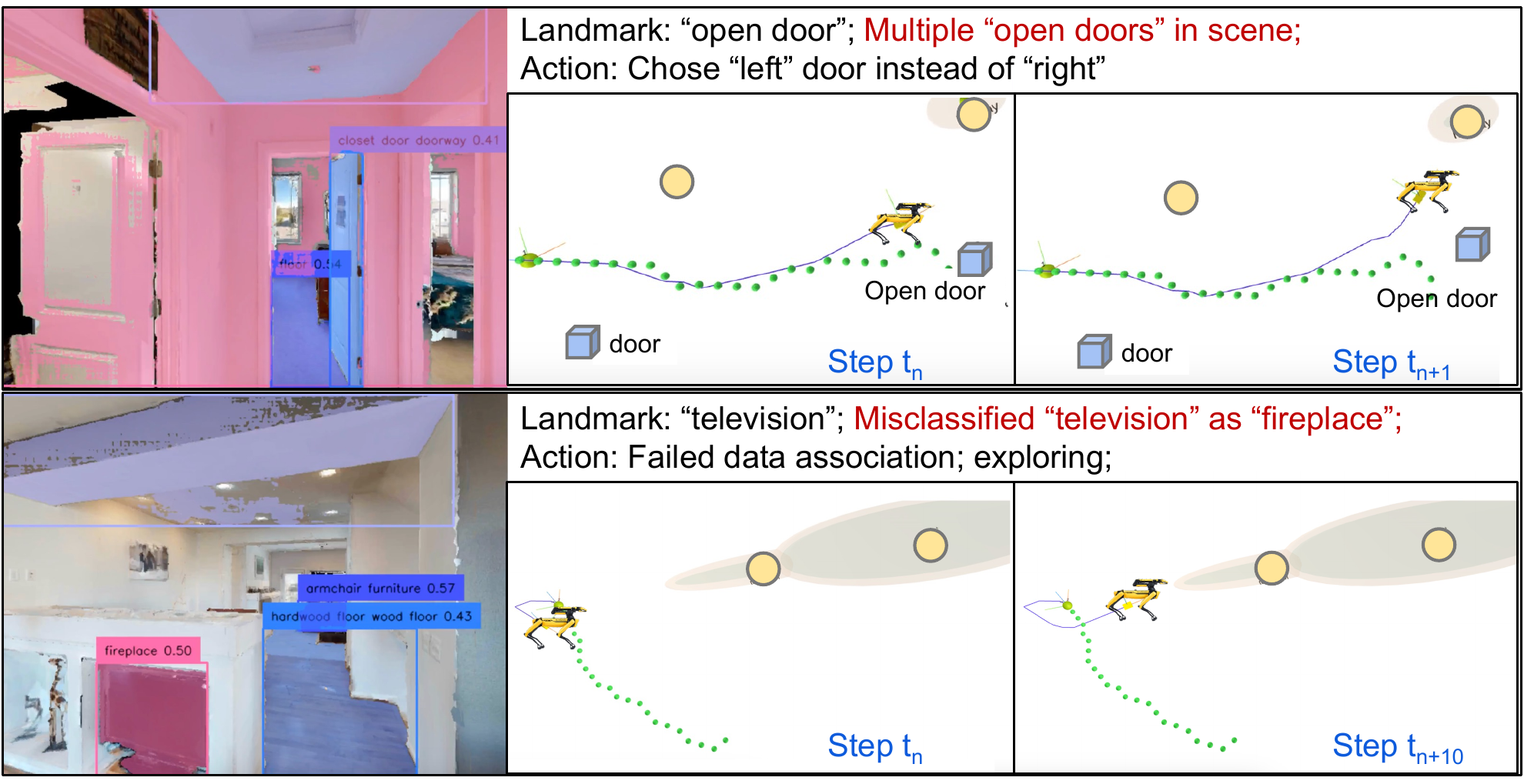}
\caption{
\textbf{Failure cases.} Two frequent failure cases in \methodname are: (top) multiple instances of the same landmark present, e.g. `door'; and (bottom) an object is mis-classified e.g. `television' identified as `fireplace'. 
}
\label{fig:error_analysis}
\vspace{-10pt}
\end{figure}

%% file: chapters/conclusion.tex
\section{Conclusion}
\label{sec:conclusion}

We introduce the \datasetnameshort dataset 
to study the distinct challenge of object-centric instruction following. 
Additionally, we propose \methodname, which robustly grounds dense language instructions in unseen visual environments, while navigating in a zero-shot manner. We achieve this via a combination of LLM-based technique and traditional navigation techniques from robotics.
Moreover, compared to prior works that use LLM as a planner, we use it much more efficiently to generate inferred graph from language instructions offline and use the widely used factor-graph based SLAM to gradually optimize the graph during navigation.
We demonstrate through experiments that \methodname outperforms two zero-shot state-of-the-art baselines across all metrics.

We find that \methodname can struggle to identify the relevant object in the presence of multiple objects of the same class. Future work in this direction could explore backtracking strategies to recover from such failures.
Actively selecting poses to obtain multiple good views of the object, perhaps drawing inspiration from active SLAM, could be another avenue for future investigation to address the other significant cause of failure in \methodname, i.e. object misclassification.

While object-centric vision-and-language navigation is far from solved as a distinct problem, future work in this area should also explore including other distinctive environmental aspects such as regions as landmarks in the instructions, making the task more challenging. 


